# Exploration of an End-to-End Automatic Number-plate Recognition neural network for Indian datasets

Sai Sirisha Nadiminti
BARC, Mumbai
nsirisha@barc.gov.in

Pranav Kant Gaur
BARC, Mumbai
pranav@barc.gov.in

Abhilash Bhardwaj
BARC, Mumbai
abhilashb@barc.gov.in

*Abstract*— Indian vehicle number plates have wide variety in terms of size, font, script and shape. Development of Automatic Number Plate Recognition (ANPR) solutions is therefore challenging, necessitating a diverse dataset to serve as a collection of examples. However, a comprehensive dataset of Indian scenario is missing, thereby, hampering the progress towards publicly available and reproducible ANPR solutions. Many countries have invested efforts to develop comprehensive ANPR datasets like Chinese City Parking Dataset (CCPD) for China [1] and Application-oriented License Plate (AOLP) dataset [2] for US. In this work, we release an expanding dataset presently consisting of 1.5k images [3] and a scalable and reproducible procedure of enhancing this dataset towards development of ANPR solution for Indian conditions. We have leveraged this dataset to explore an End-to-End (E2E) ANPR architecture for Indian scenario which was originally proposed for Chinese Vehicle number-plate recognition based on the CCPD dataset. As we customized the architecture for our dataset, we came across insights, which we have discussed in this paper. We report the hindrances in direct reusability of the model provided by the authors of CCPD because of the extreme diversity in Indian number plates and differences in distribution with respect to the CCPD dataset. An improvement of 42.86% was observed in LP detection after aligning the characteristics of Indian dataset with Chinese dataset. In this work, we have also compared the performance of the E2E number-plate detection model with YOLOv5 model, pre-trained on COCO dataset and fine-tuned on Indian vehicle images. Given that the number Indian vehicle images used for fine-tuning the detection module and yolov5 were same, we concluded that it is more sample efficient to develop an ANPR solution for Indian conditions based on COCO dataset rather than CCPD dataset.

## I. INTRODUCTION

Vehicle plate identification and recognition is used for many applications such as travel time calculation, highway car counting, traffic violation detection, and surveillance. Automatic Number Plate Recognition or ANPR reads number plates from digital images captured by cameras. In the identification of license plates, computer vision and character recognition, as well as algorithms for license plate recognition play a critical role. As a result, they are the foundation of every ANPR scheme. According to the recent development in computer vision approaches, most of the statistical methods have been replaced by deep learning neural networks due to their high accuracy in object detection [4]. However, in spite of the outstanding achievements of deep learning techniques in ANPR, ANPR datasets with cars/vehicles and number plates annotations still have a huge demand. The training data set is responsible for the progressive performance of deep learning methods. A huge training data is necessary for better utilization of more robust network architectures along with additional layers and parameters [5].

A 2-3 staged model for ANPR consists of separate detection and recognition stages and optionally a segmentation stage [6,7]. This cascaded two-stage approach separates the two phases completely. Once the detection module is trained, it cannot be reverted based on the loss of the recognition module. The end-to-end model is helpful to tackle this issue. The RPnet model is one such end-to-end number plate recognition model [8]. An end-to-end model is more beneficial than a multistage ANPR model due to various reasons. Firstly, in a multistage approach, it is important to achieve correct results in each stage, whereas in the end-to-end approach, the bounding box predictions are also changed based on the recognition loss of the recognition module, thus training both modules at once effectively. Secondly, the license plate recognition stage exploits convolution features extracted in the license plate detection stage. This allows the models to share parameters and have fewer parameters than a typical two-stage model would require, making the end-to-end model faster and more accurate. Thirdly, operations between different stages such as extracting and resizing the LP region for recognition are always accomplished by less efficient CPU, making LP recognition slower. The end-to-end models can take advantage of the fact that license plate detection and recognition being highly correlated. This work is based on exploiting the large-scale and popular dataset CCPD and adapting it to Indian conditions as Indian dataset is very limited. In this paper, the CCPD dataset and the Indian dataset is described. Then, the E2E model is discussed with the evaluation metrics. The experiments are explained with the results. It is seen that shortage of the Indian dataset and wide variety of Indian vehicle images make the E2E model that is based on CCPD dataset impractical for Indian usage.

### A. Chinese Dataset (CCPD)

The RPnet model [8] is based on the CCPD dataset which is the largest available dataset for license plates having over 250k unique car images with detailed annotations. The images of the cars are further categorized into nine categories namely, ccpd_base, ccpd_db, ccpd_fn, ccpd_rotate, ccpd_tilt, ccpd_blur, ccpd_weather, ccpd_challenge and ccpd_np to account for various factors like weather, illumination, rotation, quality and distance of the camera from the car [1].

The annotations for CCPD dataset are very extensive and include various factors like the ratio of the bounding box to the image, the horizontal and vertical tilt degree of the license plate, the left up and bottom right coordinates, all the four vertices of the license plate, the license plate number and the brightness and blurriness factors.

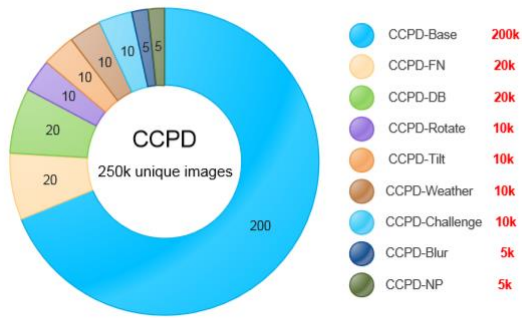

Figure 1. CCPD data distribution [8]

## B. Indian Dataset

Due to the lack of availability of any annotated public dataset for Indian vehicle images, we generated the images by web scraping from online sites like OLX and by taking vehicle images from highways at different daylight conditions. This way, we generated a dataset of about 1.5k Indian images and have made it publicly available [3]. Web scraping proved to be a valuable source to get state-wise images from OLX website. The proportion of the images for each state corresponds to the number of sellers of cars in that state, as shown in Figure 2.

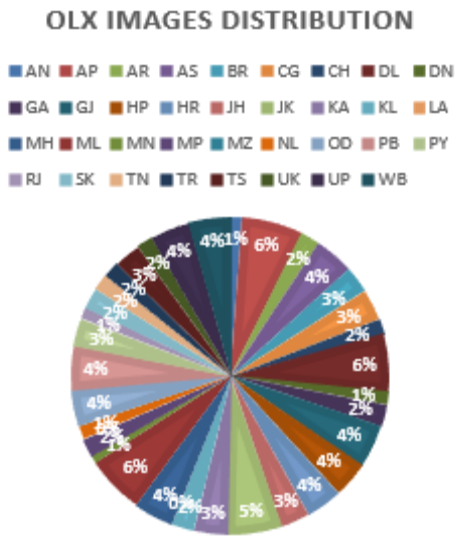

Figure 2. State-wide distribution of images obtained from OLX

Pascal VOC standard [9] was used for the manual annotation of the images, unlike the CCPD images that had the annotations in the filename of the image. We used the labelImg tool [10] for smooth annotation. Using the labelImg tool, we can automatically save the leftUp ($x_{min}$, $y_{min}$) and rightDown ($x_{max}$, $y_{max}$) coordinates as XML files in Pascal VOC format as shown in Figure 3. It is the standard format for annotating popular image datasets like ImageNet. The dataset can be expanded to include more samples subject to end-user requirement.

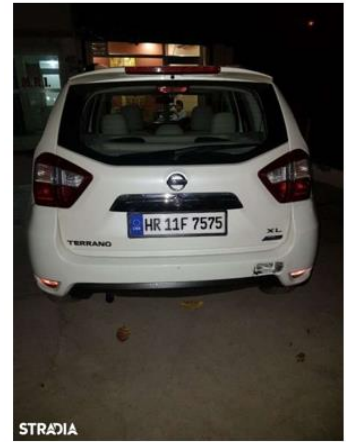

Figure 3. Example image and annotation .xml file generated by labelImg tool

## II. ALGORITHM

RPnet is composed of two modules, namely detection module (wR2) and recognition module (fh02) as shown in Figure 4.

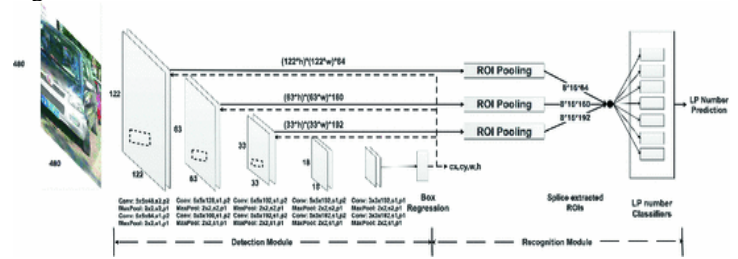

Figure 4. RPNet architecture – An end-to-end model for license plate detection and recognition [8]

### A. Detection module (wR2)

Detection module (wR2) is a deep convolutional neural network with ten convolution layers to extract different level feature maps from the input license plate image.

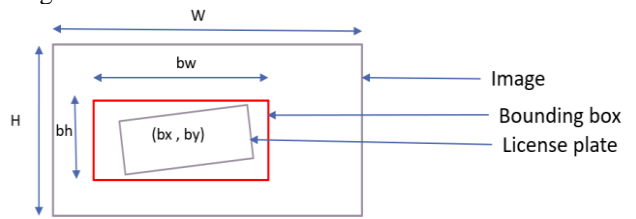

*Figure 5. Parameters to calculate the prediction of the detection module*

Before training RPnet end-to-end, the detection module must provide a reasonable bounding box prediction: ($c_x$, $c_y$, w, h) where:

$$c_x = \frac{b_x}{W}, c_y = \frac{b_y}{H}, w = \frac{b_w}{W}, h = \frac{b_h}{H}$$

Here, ($b_x$, $b_y$) depict the x and y coordinates of the center of the bounding box. ($b_w$, $b_h$) represents the width and height of the bounding box. W and H represent the width and height of the entire image. This detection module is the same for both the Chinese as well as Indian context which provides the bounding box prediction in the form of ($c_x$, $c_y$, w, h) list.

### B. Recognition module (fh02)

The recognition module exploits region-of-interest (ROI) pooling layers to extract feature maps of internet and several classifiers to predict the LP number of the LP in the input image. In Chinese license plates' context, we have seven classifiers to predict the seven characters in the license plate. For the Indian context, we have changed this to 9 classifiers for predicting 10 characters of the license plate. The first classifier is to predict the state/union territory code correctly. Each state or union territory code consists of two characters. The remaining eight characters are alphanumeric characters that may consist of the twenty-four English alphabet characters (A-Z excluding 'I' and 'O') and ten decimals (0-9).

The entire model is a single, unified network for LP detection and recognition. RPnet is trained end-to-end on CCPD and accomplishes LP bounding box detection and LP number recognition in a single forward.

### III. EVALUATION METRICS

The training involves choosing suitable loss functions for detection performance and recognition performance, as well as pre-training the detection module before training RPnet end-to-end. The localization loss is given by a smooth L1 loss and the classification loss is a cross-entropy loss. The literature [8] claims that both detection and recognition performance can be enhanced by jointly optimizing these two losses. Training objective is to minimize the combination of the detection and recognition loss.

### A. Detection accuracy metric

The bounding box is considered to be correct if and only if its IoU with the ground-truth bounding box is more than 70% (IoU > 0.7). The detection accuracy is claimed to be as high as 99.3% for ccpd_base images and as low as 84.1% for ccpd_weather images after using the wR2 module [8].

### B. Recognition accuracy metric

LP recognition is correct if and only if the IoU is greater than 0.6 and all characters in the LP number are correctly recognized. The recognition accuracy is claimed to be as high as 98.5% for ccpd_base images and as low as 85.1% for ccpd_challenge images with 61 FPS recognition rate [8].

### IV. HARDWARE AND SOFTWARE REQUIREMENTS

According to the paper [8], the wR2 and RPnet models were trained on a GPU server with 8 CPU (Intel(R) Xeon(R) CPU E5-2682 v4 @ 2.50GHz), 60GB RAM and one Nvidia GPU (Tesla P100 PCIe 16GB). All their evaluation tasks were on desktop PCs with eight 3.40 GHz Intel Core i7-6700 CPU, 24GB RAM and one Quadro P4000 GPU. In this work, Quadro M4000 GPU was used. For this work, Python 3.8.8 was used along with PyTorch 1.9.0 for Cuda 10.2.

### V. EXPERIMENTS

The objective of experimentation was to reproduce the accuracy reported in the literature [8] and attempt to customize the model and the dataset for Indian vehicles.

### A. Running the demo code for the test images

We started by running the demo code for the given sample test-data shared by CCPD authors with the pre-trained weights [1]. License plate recognition worked on 80% of test-images.

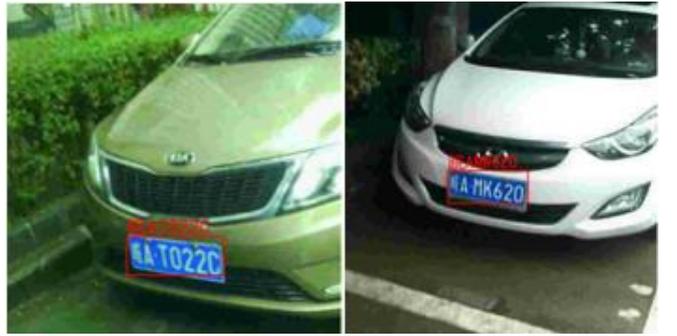

*Figure 6. Correct and incorrect predictions on the demo images*

### B. Testing the pretrained model on Indian images

The pre-trained model of RPnet was tested on the Indian images. Approximately, 10% of the images were taken as test-data. It was found that the bounding box prediction was wrong on all test-images, an example is shown in Figure 7.

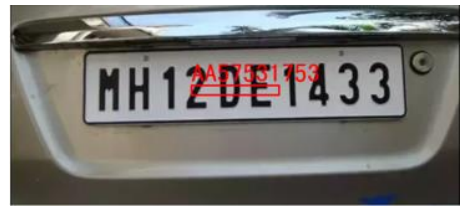

*Figure 7. Incorrect bounding box prediction on an Indian vehicle image*

### C. Inference over CCPD test images

In order to check if the model matches the accuracy that is claimed by the paper, we ran the model with all the test images of CCPD. The accuracy came out to be 41.57% for ccpd_base, 38.58% for ccpd_weather and 29.17% for ccpd_challenge. This may be due to the reason that the model that they have provided in the GitHub repository may not be the one that they have used to obtain results in the paper.

The detection module was trained using randomly taken 200K CCPD images [1]. The model generated improved the accuracy of ccpd_weather images by 20% but had no significant effect on Indian images, as shown in Table 1.

TABLE I. COMPARISION OF PRETRAINED MODEL AND OUR MODEL

| Dataset | Pre-trained model accuracy (%) | Our model accuracy (%) |
|---|---|---|
| CCPD_Weather | 61 | 80 |
| Indian images | 3 | 4.5 |

### D. Fine-tuning RPNet on Indian images

As there was no noticeable improvement on Indian images, we fine-tuned RPnet using the 25 % of Indian images over which the pre-trained detection model performed well. The experiment did not improve the accuracy of the detection model. This motivated us to align characteristics of Indian dataset with CCPD through image-preprocessing.

### E. Resizing the Indian images to 720x1160 pixels

From the original implementation [1], it was evident that the input shape for the network is 720x1160 pixels. The images were resized to 480x480 pixels. It then normalized the bounding box with respect to the original image, but used the resized images. Based on this observation, we decided to resize the images directly to 720x1160 pixels as a data preprocessing step and changed the annotations accordingly. But there was not much improvement in the bounding box prediction with the pre-trained wR2 weights because, it apparently, lead to distortion of input-features as can be seen in Figure 8.

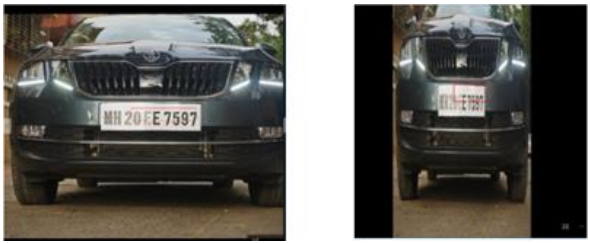

*Figure 8. No improvement after resizing the image to 720x1160 and using pre-trained wR2*

### F. Letterboxing images

Letterboxing the images resizes the images by preserving the aspect ratio and the rest of the image is padded to form the final image of size 720x1160, it eliminated the image-distortions reported above, as can be visually observed in Figure 9. This resulted in a 3.5 % improvement over original images using pre-trained weights.

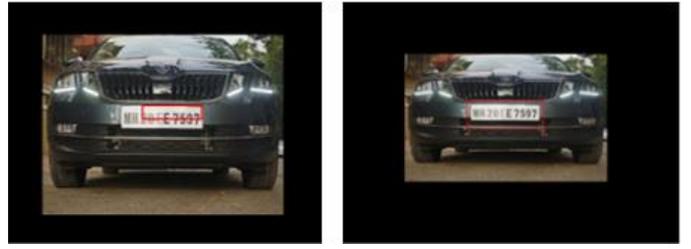

*Figure 9. Letterboxing images gave us a significant improvement on license plate detection*

### G. Fine-tuning wR2 with images having IoU>0.5

Fine-tuning wR2 with 20% images having IoU>0.5 after letterboxing increased the IoU accuracy from 6.40% to 7.55%.

### H. Data Analysis of CCPD and Indian Images

Poor generalization over Indian images motivated us to do a thorough data analysis over CCPD and Indian images. It was observed that the training and testing images of CCPD follow similar distribution. Also, the variance of the bounding box coordinates was less in case of CCPD images when compared to the Indian dataset, as shown in Table II.

TABLE II. INDIAN IMAGES DO NOT FOLLOW THE DISTRIBUTION OF CCPD IMAGES

| Dataset | $X_{min}$ | $X_{max}$ | $Y_{min}$ | $Y_{max}$ |
|---|---|---|---|---|
| CCPD_Base | 263±61 | 449±59 | 478±63 | 546±64 |
| CCPD_Weather | 227±67 | 493±63 | 474±63 | 575±65 |
| Indian images | 240±110 | 474±113 | 611±88 | 679±81 |

### I. Checking IoU after shifting the images

Indian images were preprocessed to match the CCPD distribution. It consisted of shifting the image to appropriate ($x_{min}$, $y_{min}$) bounding box location and stretching the image. We observed that IoU increased significantly after doing this, as can be seen in Figure 10 and Figure 11. We automated this process to do this translation for given set of images and also calculate the new annotations for shifted images and store them in Pascal VOC format.

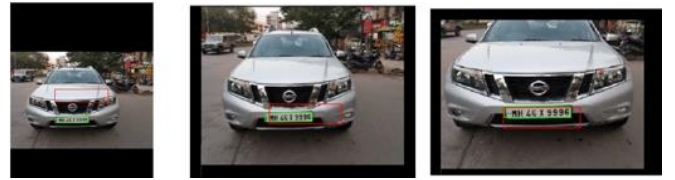

*Figure 10(a) Original image with IoU = 0.0 (b) Cropped image to reduce y coordinates to get IoU=0.28 (c) Stretched the image to accommodate cx, cy to get IoU=0.40. Green represents ground truth bounding box and red represents the predicted bounding box.*

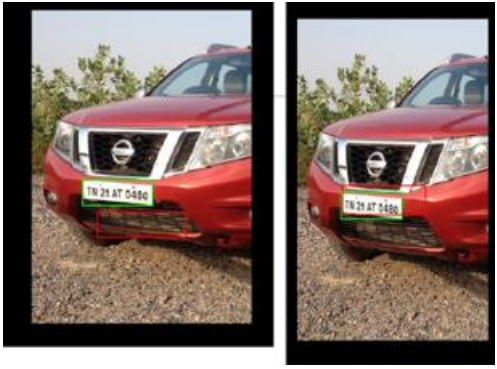

*Figure 11. Improvement of IoU after the images are shifted. Green represents ground truth bounding box and red represents the predicted bounding box*

### J. Training RPNet after adding classifiers

Fine-tuned wR2 showed better IoU. So, we used this model to train RPnet with 75% images that gave IoU>0.5 and had 10 digit characters only in their license plates. The Indian license plate for 10 characters consists of the expression in which the first two alphabets represent the state code. The nine classifiers were built accordingly in which the first classifier was based on the state. It was seen that the total loss went down by 68.54%

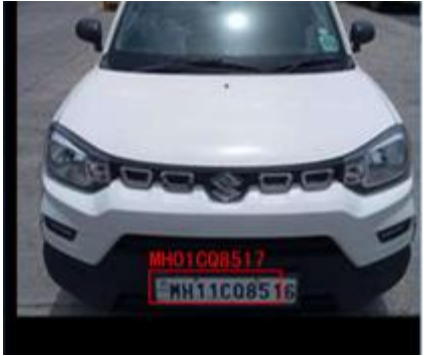

*Figure 12. The model's prediction on running over an image from train set*

### K. Using YOLOv5 for License Plate Detection

As RPnet was not able to generalize over Indian images, we attempted fine-tuning of Yolov5s (pre-trained on COCO dataset) [9] for number plate detection. The LP detection accuracy reached 99.52% after fine-tuning Yolov5s over Indian images. An example is shown in Figure 13. This experiment stated that due to the variety in Indian images, an extensive dataset like COCO is required for training.

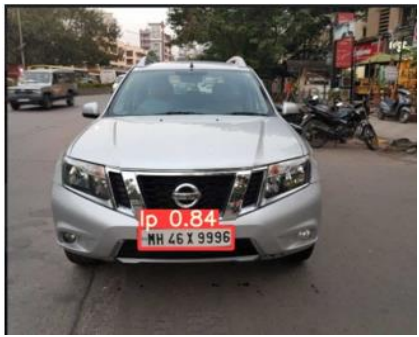

*Figure 13. Result of training YOLOv5 for license plate detection*

## VI. DISCUSSIONS

We tried to understand the reasons why CCPD trained RPnet model cannot be used as it is for Indian context. As found in the data analysis done in experiment H., the CCPD dataset is very uniform with respect to the license plates. All 250k license plate images had blue background color with white text and an identical font.

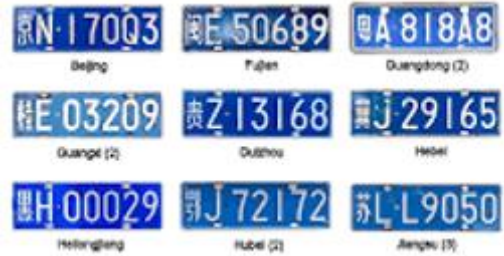

*Figure 14. Chinese license plates in CCPD are uniform*

The first problem is that the vast variety of vehicles present on a typical Indian road like cars, motorcycles, scooters, lorries, buses, auto rickshaws, SUVs, mini-trucks, vans, tractors and a majority of them having a different format and style of license plates. For example, 95% of two wheelers and auto rickshaw vehicles have multiline number plates, whereas 90% of the cars have single line plates [14]. Secondly, there is the usage of several types of fonts and custom designed plates. The plates are also of different shapes and sizes i.e., all plates are not rectangular; some are trapezoid and other irregular shapes. There are extra characters or shapes like '-' or '.' or simply large spaces in the number plate apart from the license number of the vehicle. The characters on the Indian license plates are not limited to the traditional alphanumeric dataset, but also native and indigenous letters. This is because the enforcement of a vehicle number-plate standard is still in progress in India. This creates a huge challenge, as there is no appropriate dataset that caters to this variety.

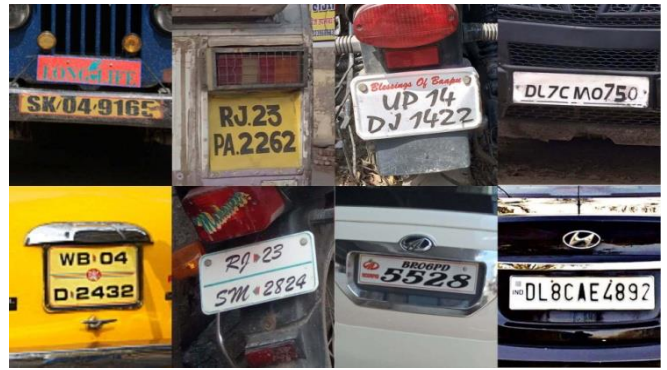

*Figure 15. Diversity in Indian license plate [13]*

The lack of large dataset persisted in our project, too. The CCPD dataset consisted of 250K images. The RPnet model was trained on 100K images and tested on the remaining images. On the contrary, we worked on 1.5K images that were too diverse in the distribution.

However, it is observed that the license plate detection improves when the Indian dataset is preprocessed in a similar way as the Chinese dataset. The preprocessing includes letterboxing and translation of the images to mimic Chinese images. This observation further proves that the Chinese images have less diversity in terms of the size of license plate in the images and the range of the coordinates of the bounding box where the license plate is predicted, due to which RPnet model did not work for the Indian vehicle images.

## VII. CONCLUSION

As the Indian dataset was sparse, we created a dataset of around 1.5k images that were annotated in Pascal VOC standard. The accuracy claimed in the literature [8] was reproduced on ccpd_weather test images with an accuracy of 80% as compared to 84% that was claimed by the paper. The generalization of detection module was improved on Indian images manifolds (from IoU=0.01 to IoU=0.325) after performing certain image preprocessing steps, although the recognition module could not be generalized on Indian test images. The reasons may be attributed to the fact that CCPD model was trained on 100K images whereas Indian images were limited. We tried an alternative approach that used Yolov5s achieving 99.5% accuracy for detection.

### ACKNOWLEDGEMENT
Authors would like to thank Computer Division, BARC for providing infrastructure and necessary support for this project. We would also like to thank Mr. Dinesh Sarode, Mr. PPK Venkat, Mr. Pritam Shete and Mrs. Mohini for guidance.

## REFERENCES

[1] Codebase and CCPD database for RPNet. [Online] https://github.com/detectRecog/CCPD.
[2] Application-oriented license plate recognition. Hsu, Gee-Sern, Chen, Jiun-Chang and Chung, Yu-Zu. s.l. : IEEE transactions on vehicular technology, 2012.
[3] Indian Vehicle Dataset. Amazon drive link. [Online] https://www.amazon.com/clouddrive/share/rrCCyD1yoEt6WqHRcprU0C1FzMFJo26xajwC9nZEEgl.
[4] Automated License Plate Recognition: A Survey on Methods and Techniques. J., Shashirangana, et al. IEEE Access, Vol. 9, pp. 11203-11225.
[5] Lubna, Mufti, Naveed and Shah, Syed Afaq Ali. Automatic Number Plate Recognition:A Detailed Survey of Relevant Algorithms. s.l. : Sensors, 2021.
[6] Real-time Bhutanese license plate localization using YOLO. Yonten, Jamtsho, Panomkhawn, Riyamongkol and Rattapoom, Waranusast. 2, s.l. : ICT Express, Vol. 6.
[7] NTUT ML license plate recognition. GitHub Link. [Online] https://github.com/hsuRush/DeepANPR.
[8] Towards End-to-End License Plate Detection and Recognition: A Large Dataset and Baseline. Xu, Zhenbo, et al. s.l. : Springer, 2018. ECCV.
[9] Everingham, Mark and Winn, John. The PASCAL Visual Object Classes Challenge 2012 (VOC2012) Development Kit. [Online] May 2012. https://pjreddie.com/media/files/VOC2012_doc.pdf.
[10] LabelImg tool GitHub Link. [Online] https://github.com/tzutalin/labelImg.
[11] Yolov5 GitHub Link. [Online] https://github.com/ultralytics/yolov5.
[12] Automatic License Plate Recognition for Indian Roads Using Faster-RCNN. P., Ravirathinam and A., Patawari. 2019. 11th International Conference on Advanced Computing (ICoAC).
[13] ANPR for India. [Online] https://platerecognizer.com/anpr-for-india/.
[14] SLPNet: Towards End-to-End Car License Plate Detection and Recognition Using Lightweight CNN. Zhang, W., Mao, Y. and Han, Y. s.l. : Pattern Recognition and Computer Vision, 2020.